\documentclass{article}

\PassOptionsToPackage{numbers, compress}{natbib}

\usepackage[preprint]{neurips_2025}
\usepackage{graphicx}
\usepackage{booktabs}
\usepackage{subcaption}
\usepackage[utf8]{inputenc}
\usepackage[T1]{fontenc}
\usepackage{hyperref}
\usepackage{url}
\usepackage{amsfonts}
\usepackage{amsmath}
\usepackage{nicefrac}
\usepackage{microtype}
\usepackage{xcolor}
\usepackage{newunicodechar}
\newunicodechar{≤}{\ensuremath{\le}}
\newunicodechar{≥}{\ensuremath{\ge}}
\newunicodechar{≈}{\approx}

\title{SafeBench-Seq: A Homology-Clustered, CPU-Only Baseline for Protein Hazard Screening with Physicochemical/Composition Features and Cluster-Aware Confidence Intervals}

\author{%
  Muhammad Haris Khan\\
  University of Copenhagen\\
  \texttt{muhammad.kahn@di.ku.dk} \\
}

\begin{document}

\maketitle

\begin{abstract}
Foundation models for protein design raise concrete biosecurity risks, yet the community lacks a simple, reproducible baseline for sequence-level hazard screening that is explicitly evaluated under homology control and runs on commodity CPUs. We introduce \emph{SafeBench-Seq}, a metadata-only, reproducible benchmark and baseline classifier built entirely from public data (SafeProtein hazards and UniProt benigns) and interpretable features (global physicochemical descriptors and amino-acid composition). To approximate “never-before-seen” threats, we homology-cluster the combined dataset at ≤40\% identity and perform \emph{cluster-level holdouts} (no cluster overlap between train/test). We report discrimination (AUROC/AUPRC) and screening-operating points (TPR@1\%~FPR; FPR@95\%~TPR) with 95\% bootstrap confidence intervals (n=200), and we provide calibrated probabilities via \texttt{CalibratedClassifierCV} (isotonic for Logistic Regression / Random Forest; Platt sigmoid for Linear SVM). We quantify probability quality using Brier score, Expected Calibration Error (ECE; 15 bins), and reliability diagrams. Shortcut susceptibility is probed via composition-preserving residue shuffles and length-/composition-only ablations. Empirically, random splits substantially overestimate robustness relative to homology-clustered evaluation; calibrated linear models exhibit comparatively good calibration, while tree ensembles retain slightly higher Brier/ECE. SafeBench-Seq is CPU-only, reproducible, and releases \emph{metadata only} (accessions, cluster IDs, split labels), enabling rigorous evaluation without distributing hazardous sequences. \noindent\textbf{Code \& metadata:} \href{https://github.com/HARISKHAN-1729/SafeBench-Seq}{https://github.com/HARISKHAN-1729/SafeBench-Seq}

\end{abstract}

\section{Introduction}

Generative AI (GenAI) for protein design have dramatically increased the ability to create novel biomolecules, but they also raise urgent biosecurity concerns. Recent work highlights that the lack of systematic “red-teaming” of protein foundation models enables potential misuse, including the de novo design of hazardous proteins \citep{Fan2025SafeProtein}, \cite{Hunter2024Biosecurity}. For example, experts warn that DNA synthesis screening – which relies on homology to known toxins – can be evaded by AI-generated “synthetic homologs” of dangerous proteins \cite{Hunter2024Biosecurity}, \citep{Ikonomova2025ProteinDesignRisk}. Efforts by Wittmann et al.\ have experimentally demonstrated that generative pipelines can reformulate toxic protein sequences to bypass standard screens \citep{Ikonomova2025ProteinDesignRisk}. These findings underscore that protein generative models already pose a biosecurity risk, making effective screening and defense strategies critical.

Despite this risk, there is a striking lack of simple, reproducible baselines for sequence-level protein hazard filtering. Existing pipelines for nucleic acid orders typically use BLAST or related alignments against curated hazard databases \citep{reference_database_hazardous_sequences}, but no lightweight open-source classifiers have been made available specifically for protein-level screening. Most prior tools focus on specialized toxin prediction (e.g., venom, peptide toxins) or require heavy GPU models. In contrast, our goal is to provide a tractable CPU-only baseline hazard detector for arbitrary protein sequences, built entirely from public data and executable on standard platforms (e.g., Colab) under strict safety constraints. We deliberately avoid releasing any novel sequences (only metadata such as accession IDs and taxonomic labels) and provide only aggregate statistics, to mitigate any risk of misuse of our benchmark data.

A key aspect of realistic hazard screening is robust evaluation. We emphasize homology-aware splits and stringent metrics to avoid overly optimistic results. In particular, we follow recommendations to cluster by sequence identity and hold out \emph{homology clusters} (≤40\% identity) to simulate “never-before-seen” threats, rather than random splits \citep{reference_database_hazardous_sequences},\citep{Zhu2025ToxDL}. We also measure performance at extreme operating points (e.g., TPR at 1\%~FPR, FPR at 95\%~TPR) to reflect the high-consequence setting where false positives must be rare. Finally, we probe models for shortcut biases: length and taxonomy can trivially distinguish many known toxins; signal peptides are left for future work.

In this work, we introduce \emph{SafeBench-Seq}, which is, to our knowledge, a tractable, CPU-only baseline evaluated under homology-clustered splits with explicit spurious-signal probes. Our baseline uses only open data (SafeProtein’s hazard sequences and UniProt) and simple ML models (logistic regression, SVM, random forest) with calibrated probability outputs. By anchoring evaluation in realistic homology splits and spurious-signal probing, we aim to set a clear baseline for future sequence-based biohazard detectors, complementing the rich set of toxin-focused classifiers and protein-design red-teaming efforts in the literature.

\section{Related Work}

Protein hazard screening has long relied on sequence alignment pipelines, such as BLAST against curated toxin/pathogen databases, as recommended by the International Gene Synthesis Consortium (IGSC) \citep{reference_database_hazardous_sequences}. While effective for known sequences, such methods are vulnerable to “synthetic homologs” that evade strict identity thresholds \citep{reference_database_hazardous_sequences,Hunter2024Biosecurity}.  

To address these gaps, machine learning approaches have emerged. Early classifiers like ToxinPred \citep{sharma2022toxinpred2} and ClanTox \citep{naamati2009clantox} used curated motifs and physico-chemical descriptors. More recent methods leverage deep learning and large-scale data, e.g., TOXIFY \citep{Dobson2019TOXIFY} on venom proteins, MultiToxPred \citep{Beltran2024MultiToxPred} with gradient-boosted ensembles over multiple toxin classes, and ToxDL 2.0 \citep{Zhu2025ToxDL} integrating pretrained LMs with structure-aware features. While powerful, many prior evaluations rely on random splits and GPU-heavy stacks, which may overestimate robustness against novel hazards.  

In parallel, benchmarks like TAPE \citep{NEURIPS2019_37f65c06} and large models such as ESM \citep{doi:10.1073/pnas.2016239118} have transformed protein ML, enabling diffusion-based design \citep{zhang2024foldmarkprotectingproteingenerative} and raising security concerns. Commentaries warn that GenAI-driven protein design may render homology-based screening obsolete \citep{Hunter2024Biosecurity}, prompting proposals for artificial sequence databases as defensive infrastructure \citep{Hunter2024Biosecurity,Baker2024Science}. Complementary efforts include watermarking \citep{zhang2024foldmarkprotectingproteingenerative,Chen2025Watermarking} and unlearning \citep{Li2024.10.02.616274}. Our contribution is a simple, interpretable, CPU-executable baseline evaluated under homology control, aligned with the SafeProtein ethos \citep{reference_database_hazardous_sequences,Zhu2025ToxDL}.

\section{Methods}

\subsection*{Data Curation and Filtering}
\textbf{Positives (hazards).} We use \emph{protein toxins} from SafeProtein \citep{Fan2025SafeProtein}, explicitly excluding viral proteins to focus the positive class on toxins.  
\textbf{Negatives (benign).} We sample UniProtKB (2024 release) entries that (i) do not contain the “Toxin [KW-0800]” keyword and (ii) are non-viral.  
\textbf{Sequence inclusion criteria.} We require only canonical amino acids and lengths in $[30, 1000]$ residues. We remove exact duplicates (100\% identity).  
\textbf{Length matching.} To blunt trivial length cues, we downsample the benign pool to match the positive class length distribution (quantile bin matching). This equalizes marginal length statistics (e.g., similar medians and ranges).  
\textbf{Safety posture.} We never release raw amino acid sequences to prevent potential misuse of hazardous proteins in our positive class. Our released artifacts contain only metadata (accession IDs, sequence lengths, labels, data sources, CD-HIT cluster IDs, and train/test split assignments) sufficient for result reproduction.

\subsection*{Homology-Clustered Splits}
We homology-cluster the \emph{combined} dataset (positives + negatives) at ≤40\% identity using CD-HIT \citep{huang2010cdhit}, so sequences placed in different clusters share $\leq 40\%$ identity. We evaluate under two protocols:

\textbf{Cluster split}—We perform cluster-level stratified sampling to ensure both hazardous and benign proteins are represented in train/test partitions. Each cluster receives a majority label based on its most frequent constituent class, then we randomly assign 80\% of clusters (477/597) to training and 20\% (120/597) to testing while maintaining stratification by majority label. No sequences from the same cluster appear in both partitions. Because cluster sizes are highly variable (ranging from 1 to $n$ sequences per cluster), this cluster-wise 80/20 split produces a sequence-wise 83/17 distribution (708 train, 146 test sequences).

\textbf{Random split}—a \emph{sequence-level} split that ignores clusters ($≈80/20$: 683 train / 171 test).

This design reduces homology leakage and shortcut learning and better reflects screening of new designs \citep{Zhu2025ToxDL}. In aggregate, SafeBench-Seq contains 854 examples (427 harmful / 427 benign) forming 597 clusters; 477 clusters (80\%) are assigned to train and 120 (20\%) to test in the cluster split.

\subsection*{Features and Models}
\textbf{Handcrafted feature blocks.}  
(1) \emph{20-aa composition:} for sequence $s$ of length $L$, composition is $c(a)=\frac{\#(a\in s)}{L}$ for $a\in\{A,C,D,\ldots,Y\}$.  
(2) \emph{Global physicochemical descriptors (ProtParam):} length $L$; molecular weight (MW); isoelectric point (pI); GRAVY (grand average hydropathy); aromaticity; instability index; an aliphatic-index heuristic; net charge at pH 7.0. The (Ikai) aliphatic index is approximated as $100\cdot(x_A + 2.9x_V + 3.1x_I + 3.9x_L)$ where $x_\cdot$ are residue fractions. These descriptors summarize coarse biochemical tendencies without learned embeddings.

\textbf{Classifiers (CPU-only).} We train L2-regularized Logistic Regression, a Linear SVM, and a Random Forest (400 trees) using scikit-learn. Linear models are wrapped in a pipeline with median imputation and standardization. Hyperparameters follow the CPU-only baseline used in our reference implementation (e.g., Logistic Regression $C{=}0.5$; Linear SVM $C{=}1.0$; RF class\_weight balanced\_subsample). Random seeds are fixed (1337) for splits, model fitting, and bootstrap.

\textbf{Probability calibration.} We apply \texttt{CalibratedClassifierCV} (5-fold on the training partition) to obtain well-behaved probabilities: \emph{isotonic} for Logistic Regression and Random Forest, and \emph{Platt} (sigmoid) for the Linear SVM. The same protocol is used for both the cluster split and the random split. All operating-point metrics use these calibrated probabilities.

\subsection*{Evaluation Metrics and Uncertainty}
\textbf{Discrimination.} AUROC and AUPRC computed on the held-out test set.  
\textbf{Operating points.} \emph{TPR@1\%FPR} (sensitivity at 1\% false-positive rate) and \emph{FPR@95\%TPR} (false-positive rate when sensitivity is 95\%). For each, we select the left-most threshold that achieves the target constraint on the empirical ROC curve.  
\textbf{Calibration.} We report \emph{Brier score} and \emph{Expected Calibration Error} (ECE) with 15 equal-width bins on $[0,1]$. Reliability diagrams plot the empirical fraction positive vs.\ predicted probability per bin, overlaid with the identity line.  
\textbf{Uncertainty quantification.} We use test-set bootstrapping ($n{=}200$ iid resamples with replacement) and report 95\% CIs via the 2.5/97.5 percentiles, skipping degenerate resamples lacking both classes. Unless noted otherwise, the cluster split is our primary evaluation, and the random split is reported for context.

\subsection*{Spurious-Signal Probes and Subgroup Analysis}
\textbf{Probes.} (i) \emph{Composition-preserving residue shuffle} on test sequences: per sequence, residues are randomly permuted (deterministic seed by accession), preserving length and overall composition but destroying motif order; (ii) \emph{length-only} and \emph{composition-only} ablations trained and evaluated under the same split protocol. Retaining high performance under these probes would indicate heavy reliance on trivial cues.  
\textbf{Subgroups.} We report AUROC/AUPRC across: (a) sequence-length bins (quantiles), (b) toxin clusters/families (for positives, when sufficient support exists), and (c) superkingdom (Bacteria/Archaea/Eukaryota) for negatives. We require sufficient sample size ($\geq$15) and label variability within a subgroup to report metrics.

\section{The SafeBench-Seq Benchmark}

\subsection*{Construction and Safety}
SafeBench-Seq draws hazards from SafeProtein-Bench (\emph{protein toxins only; viral proteins excluded}) \citep{Fan2025SafeProtein} and benigns from UniProt after strict exclusion filters (no KW-0800, non-viral). We keep length 30–1000 aa, restrict to canonical amino acids, deduplicate exact matches, and length-match benigns to positives to blunt trivial cues. The core artifact is a \emph{metadata-only} CSV listing accession, label, length, source, CD-HIT cluster ID, and split assignments (random and cluster). Users can reproduce results by fetching sequences from public sources via accession; we do not distribute raw sequences.

\subsection*{Splits and Counts}
\begin{figure}[!htb]
  \centering
  \includegraphics[width=1\linewidth]{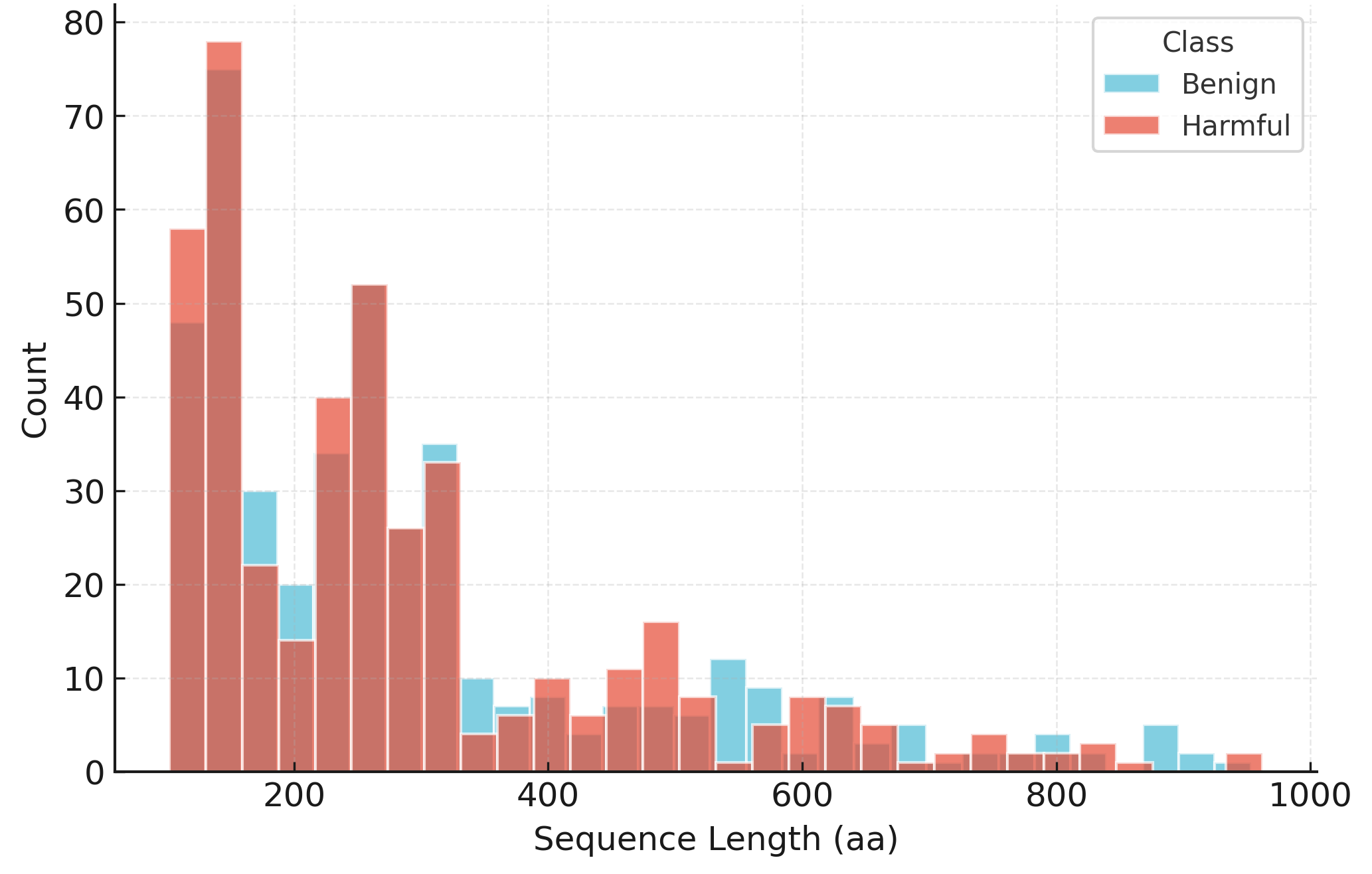}
  \caption{Sequence length distributions for harmful (red) vs.\ benign (blue-green) proteins in SafeBench-Seq. Lengths are closely matched by design, limiting trivial cues.}
  \label{fig:length-distribution}
\end{figure}

\textbf{Random split.} $\approx$80/20 by sequence: 683 train / 171 test.  
\textbf{Cluster split.} CD-HIT clustering at $≤40\%$ identity \citep{huang2010cdhit}. We hold out clusters: 477 of 597 clusters (80\%) in train; 120 (20\%) in test. Because cluster sizes vary, the \emph{sequence}-level counts are 708 train and 146 test ($≈83/17$).  
\textbf{Dataset size.} 854 sequences total, balanced (427 harmful / 427 benign).  
\textbf{Length distribution.} Classes are closely matched by design (e.g., median $≈246$ aa; mean $≈290$ aa). See Fig.~\ref{fig:length-distribution}.

\section{Results}
\begin{table*}[t]
\centering
\caption{Discrimination performance (calibrated models) on random and cluster splits using the \emph{physicochemical and composition*} feature set. Metrics with 95\% CIs from bootstrap (n=200); best values per split are in \textbf{bold}.}
\label{tab:main_results}
\resizebox{\textwidth}{!}{%
\begin{tabular}{llcccc}
\toprule
Split & Model & AUROC & AUPRC & TPR@1\%FPR $\uparrow$ & FPR@95\%TPR $\downarrow$ \\
\midrule
Random & LogReg (*) & 0.901 [0.849–0.944] & 0.894 [0.820–0.955] & 0.198 [0.134–0.470] & 0.459 [0.145–1.000] \\
       & LinSVM (*) & 0.898 [0.842–0.945] & 0.896 [0.833–0.953] & 0.151 [0.103–0.449] & 0.435 [0.141–0.946] \\
       & RF (*)     & \textbf{0.953 [0.917–0.979]} & \textbf{0.944 [0.883–0.983]} & \textbf{0.267 [0.200–0.844]} & \textbf{0.353 [0.056–0.481]} \\
\midrule
Cluster & LogReg (*) & 0.871 [0.803–0.925] & 0.851 [0.772–0.917] & 0.095 [0.050–0.272] & 0.601 [0.250–1.000] \\
        & LinSVM (*) & 0.865 [0.798–0.918] & 0.842 [0.768–0.910] & 0.083 [0.047–0.245] & 0.629 [0.262–0.983] \\
        & RF (*)     & \textbf{0.919 [0.866–0.961]} & \textbf{0.905 [0.832–0.964]} & \textbf{0.121 [0.067–0.296]} & \textbf{0.512 [0.183–0.892]} \\
\bottomrule
\end{tabular}}
\end{table*}

Table~\ref{tab:main_results} summarizes discrimination performance for three \emph{calibrated} classifiers (Logistic Regression, Linear SVM with Platt, Random Forest with isotonic) using the \emph{base} feature set on both splits. We report AUROC, AUPRC, and screening-operating points, with 95\% bootstrap CIs (n=200). Random-split performance is high (e.g., RF AUROC $\approx$0.95), but the cluster split reduces AUROC by roughly 4–8 points and sharply depresses tail performance (TPR@1\%FPR drops from $\approx$0.27 to $\approx$0.12). This gap indicates that homology-aware evaluation better reflects the challenge of screening \emph{novel} homology groups and that random splits overestimate robustness. Across both splits, Random Forest leads the linear baselines, though margins narrow under the cluster split. Operating-point metrics show wider CIs as expected for imbalanced data and extreme thresholds.

\subsection{Calibration}

\begin{figure}[t]
\centering
\includegraphics[width=0.32\linewidth]{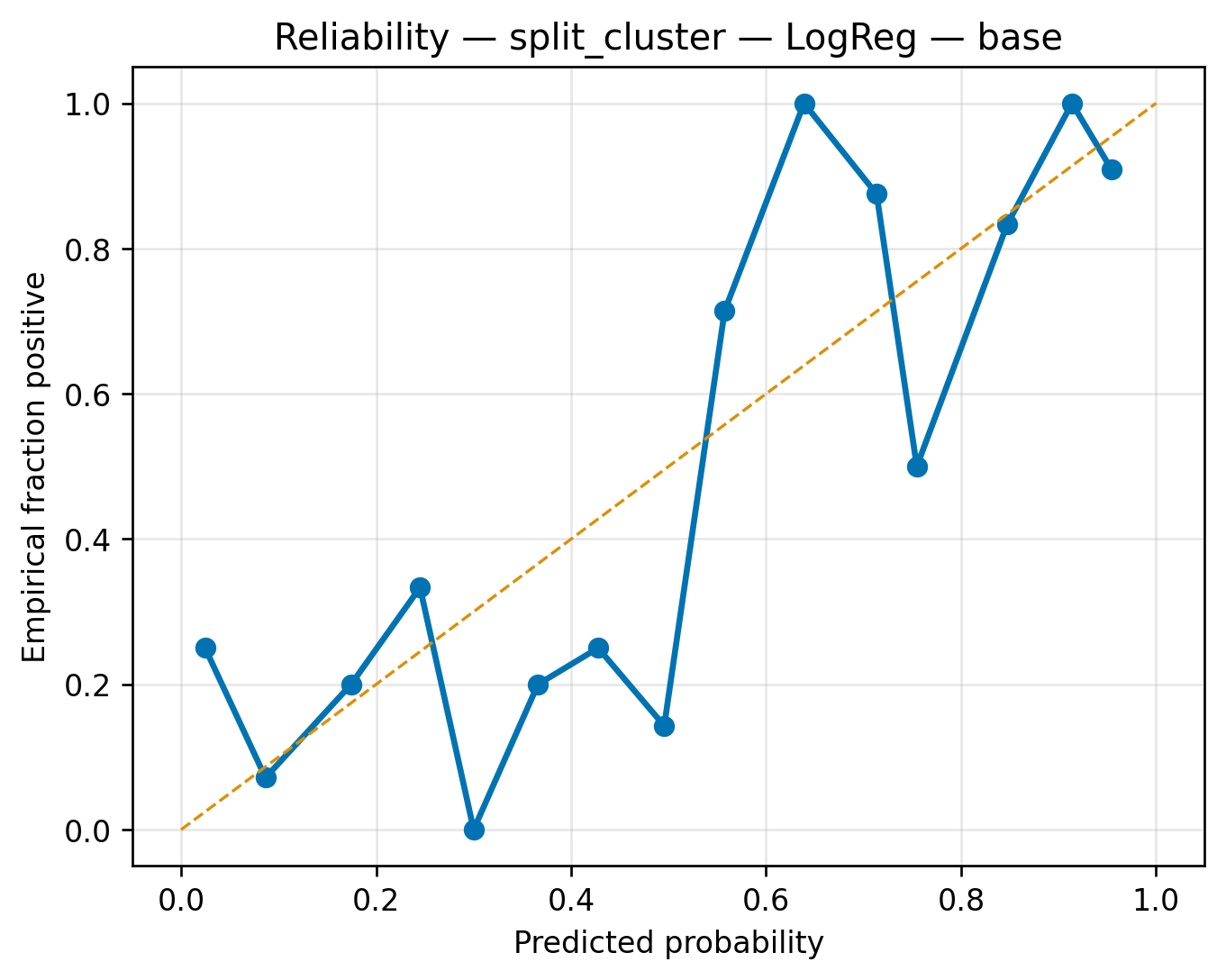}
\includegraphics[width=0.32\linewidth]{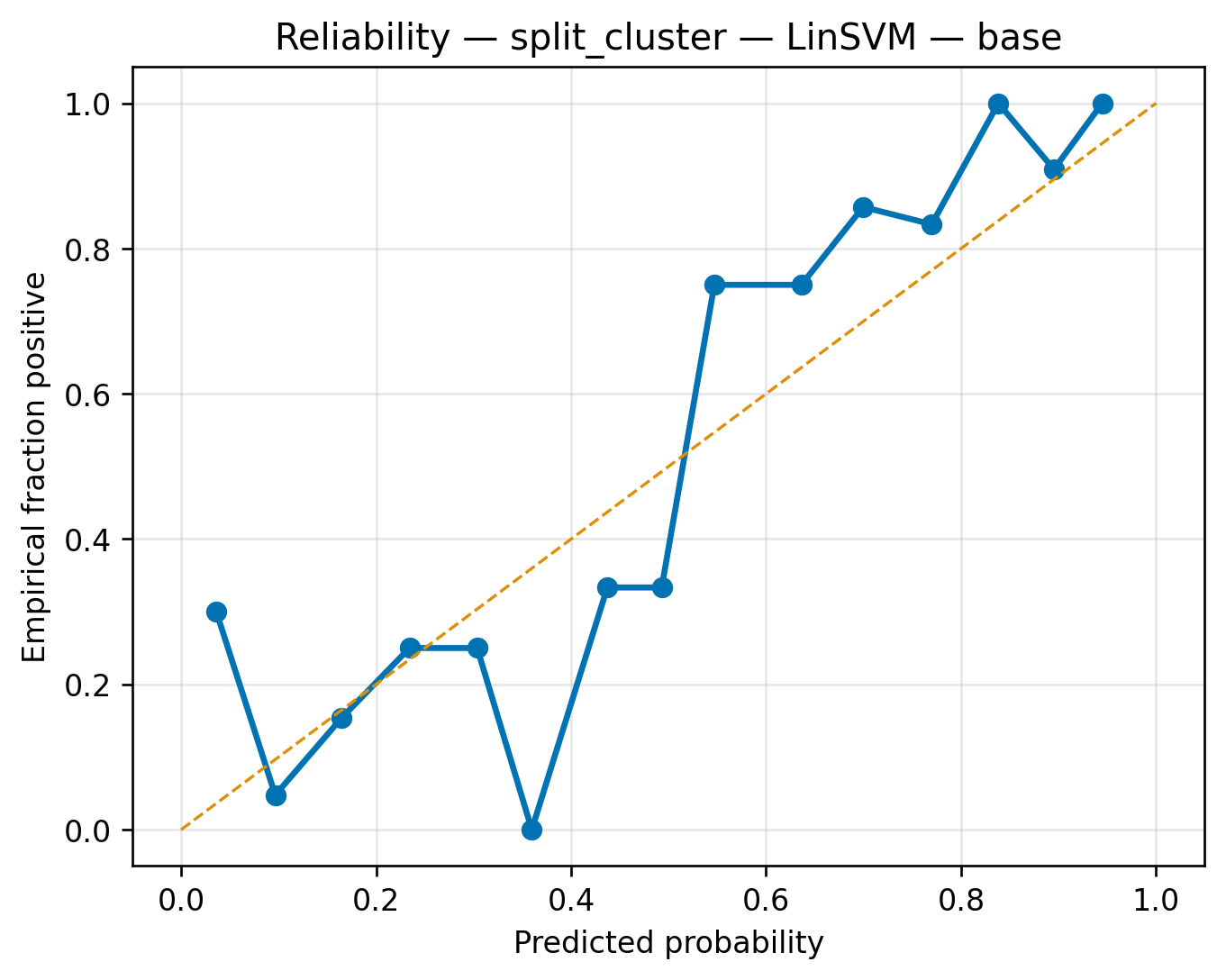}
\includegraphics[width=0.32\linewidth]{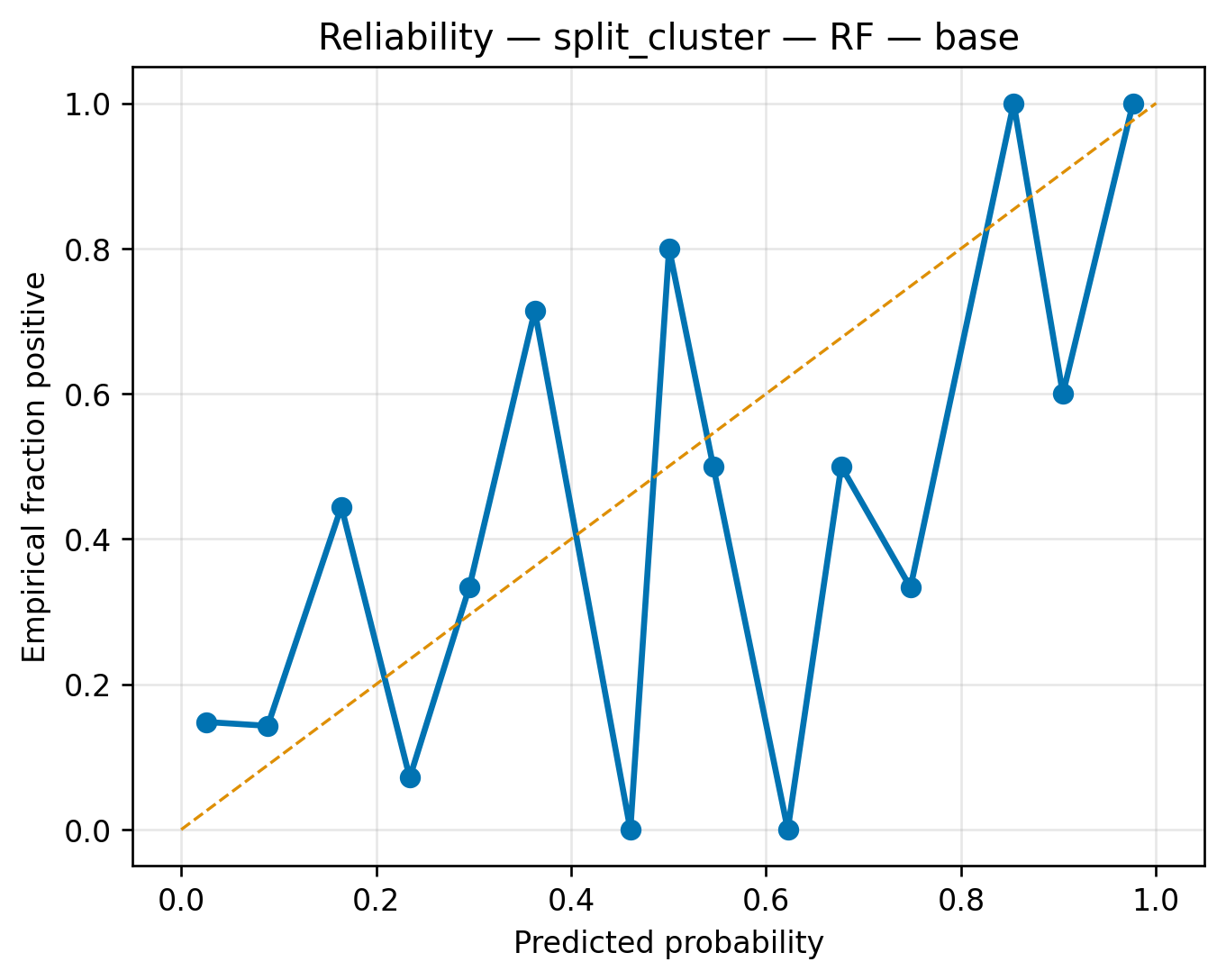}
\caption{Reliability diagrams on the cluster split (base features). Dashed line: perfect calibration.}
\label{fig:reliability_cluster}
\end{figure}

We calibrate probabilities with \texttt{CalibratedClassifierCV} (isotonic for LogReg/RF; Platt for LinSVM). Beyond discrimination, Table~\ref{tab:calibration_cluster} reports Brier and ECE on the cluster split with \emph{base} features; Fig.~\ref{fig:reliability_cluster} shows reliability diagrams. Linear models are comparatively well-calibrated post hoc; Random Forest exhibits slightly higher residual miscalibration (higher Brier/ECE), though calibration still reduces deviations from the diagonal.

\begin{table*}[t]
\centering
\caption{Calibration metrics on the cluster split (base features). 95\% CIs via bootstrap (n=200). AUROC/AUPRC appear in Table~\ref{tab:main_results}.}
\label{tab:calibration_cluster}
\setlength{\tabcolsep}{6pt}\scriptsize
\begin{tabular}{lcc}
\toprule
Model & Brier $\downarrow$ & ECE $\downarrow$ \\
\midrule
LinSVM & 0.143 [0.112, 0.181] & 0.111 [0.108, 0.195] \\
LogReg & 0.140 [0.104, 0.183] & 0.118 [0.106, 0.192] \\
RF     & 0.156 [0.115, 0.204] & 0.138 [0.105, 0.204] \\
\bottomrule
\end{tabular}
\end{table*}

\subsection{Spurious-Signal Probes}

Length-only and composition-only features yield near-random AUROC ($≈0.50–0.55$), indicating trivial cues were largely equalized. A composition-preserving residue shuffle of test sequences (randomizing residue order per sequence) degrades performance further, showing that models leverage motif/ordering information rather than bulk composition alone. These probes support that the screen captures sequence-level hazard signals, not dataset artifacts.

\subsection{Subgroup Performance}

\begin{figure}[t!]
  \centering
  \includegraphics[width=0.48\linewidth]{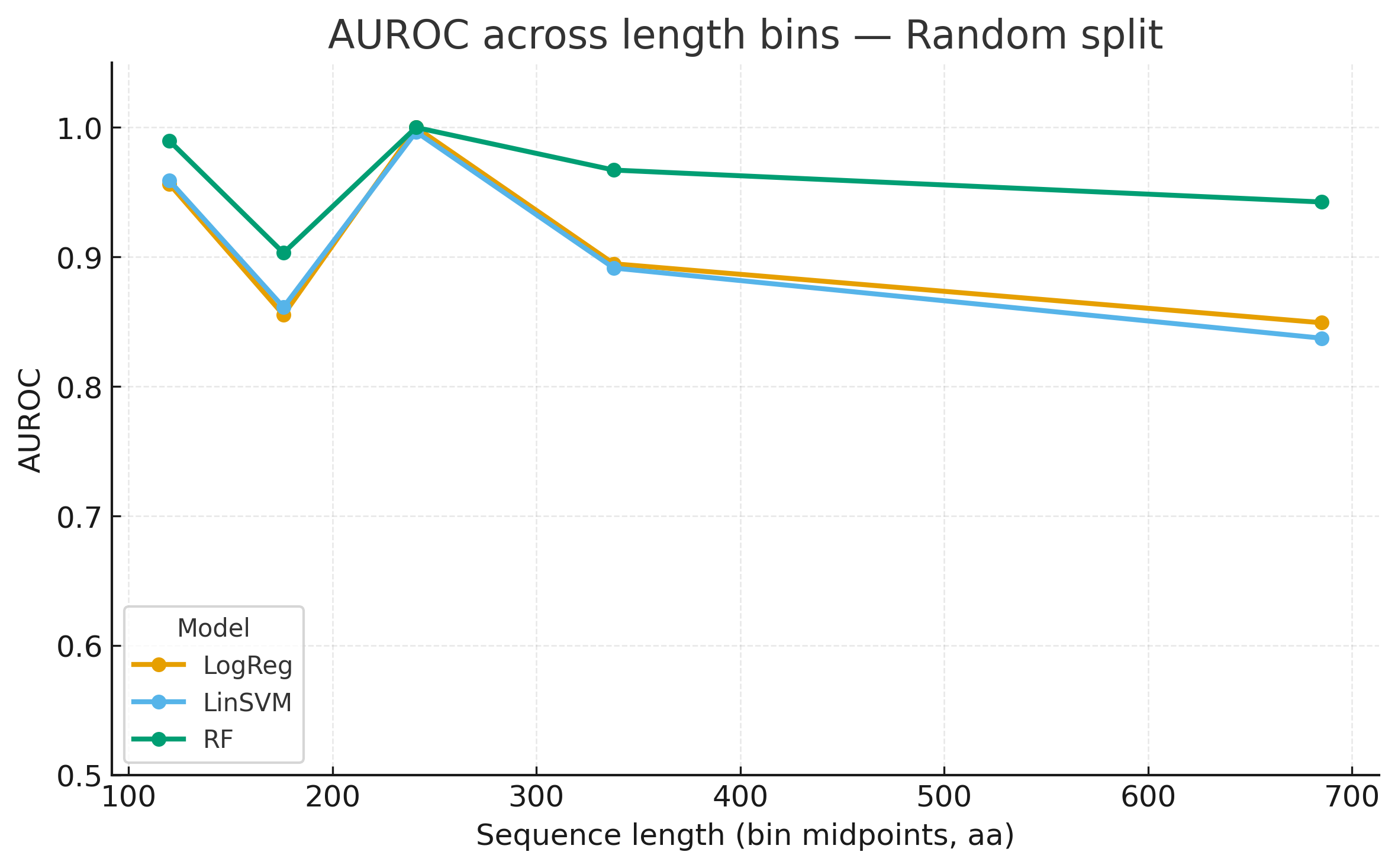}\hfill
  \includegraphics[width=0.48\linewidth]{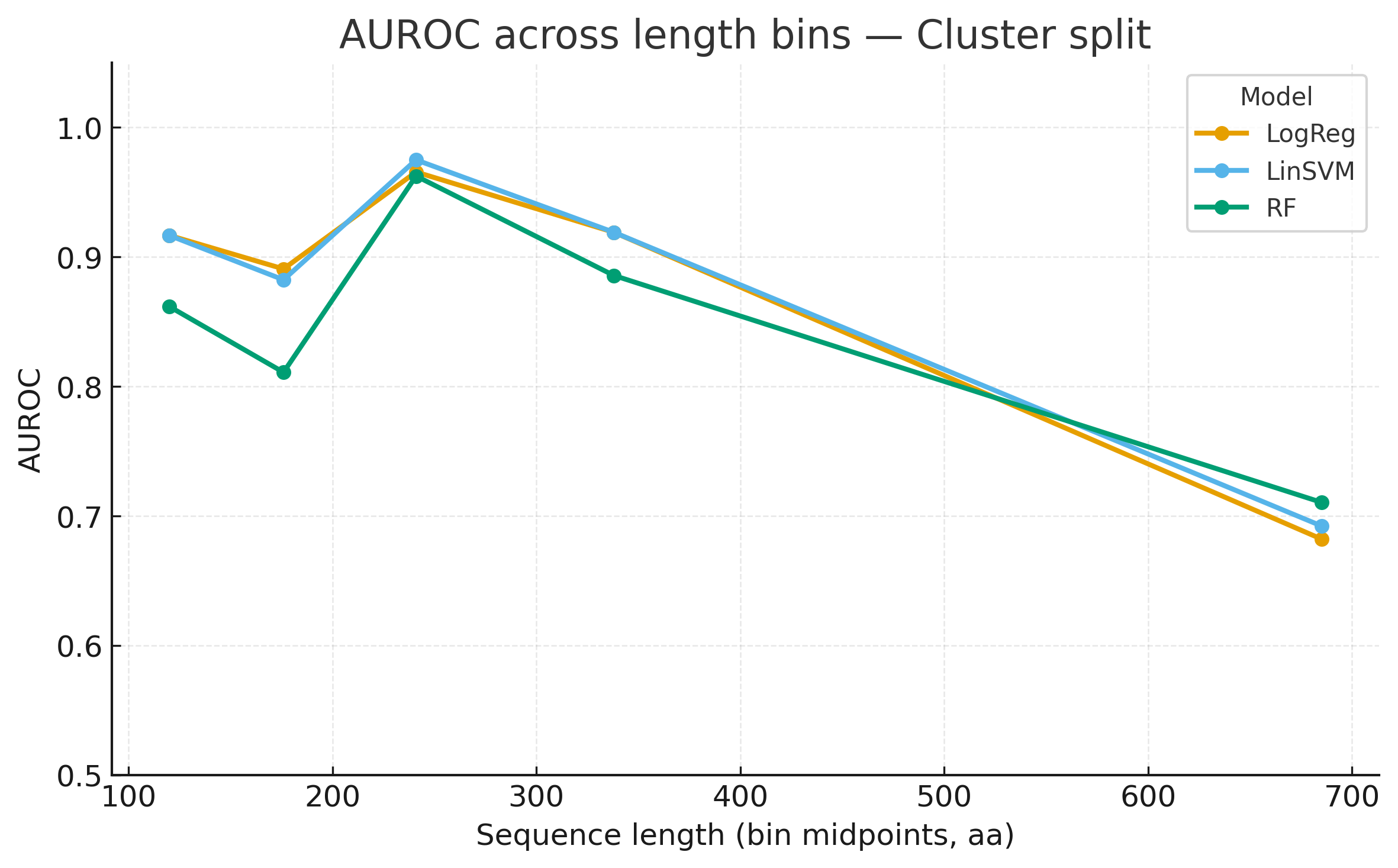}
  \caption{AUROC across sequence-length bins (left: random split; right: cluster split). Mid-lengths are easiest; very long proteins degrade under homology control.}
  \label{fig:len_bins_auroc}
\end{figure}

\begin{figure}[t!]
  \centering
  \includegraphics[width=0.48\linewidth]{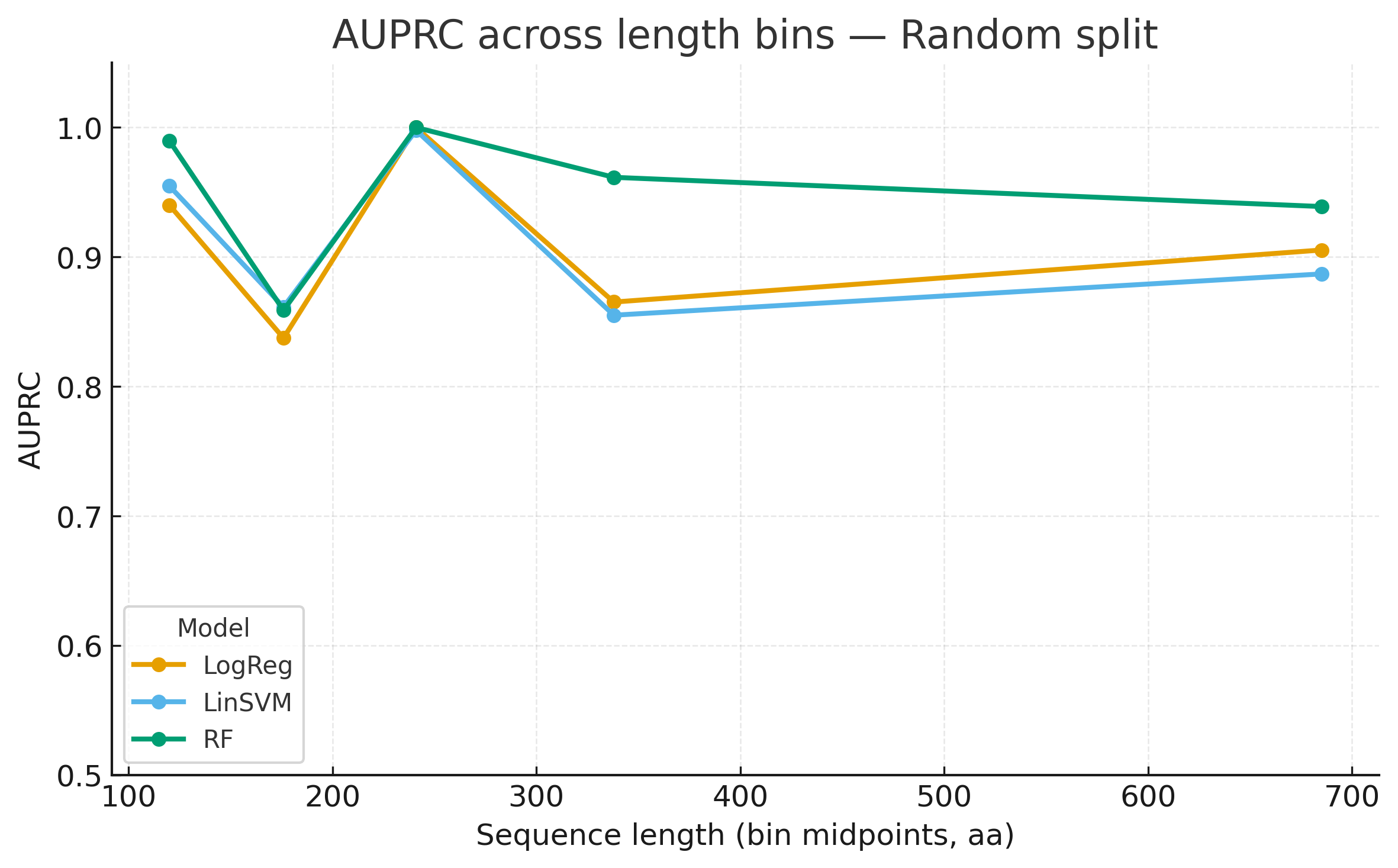}\hfill
  \includegraphics[width=0.48\linewidth]{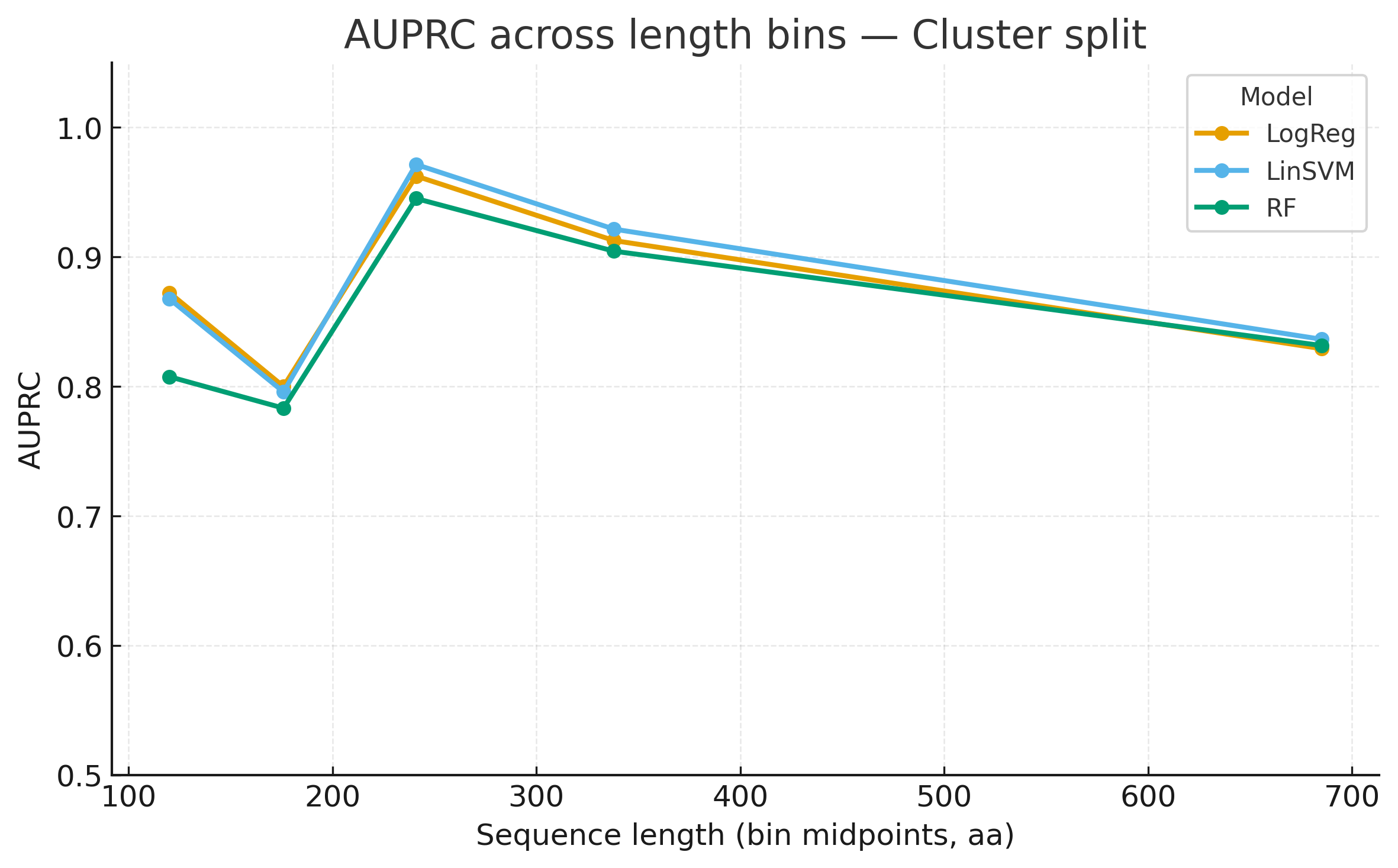}
  \caption{AUPRC across sequence-length bins mirrors AUROC: mid-lengths peak; long proteins suffer under the cluster split.}
  \label{fig:len_bins_auprc}
\end{figure}

We break down performance by sequence length, toxin family (positives), and superkingdom (negatives). Fig.~\ref{fig:len_bins_auroc}–\ref{fig:len_bins_auprc} show AUROC/AUPRC across length bins: performance peaks for mid-length proteins (200–300 aa) and degrades for very long sequences ($>$400 aa), especially under the cluster split—consistent with reduced generalization to multi-domain proteins. Family- and taxonomy-level results (Appendix) show heterogeneity: some toxin clusters are near-perfectly separated while others cluster around AUROC $\approx$0.80; bacterial negatives are generally easier than eukaryotic ones.

\paragraph{Takeaways.}
(1) Homology-clustered evaluation surfaces a substantial robustness gap vs.\ random splits—critical for screening novel sequences. (2) Post-hoc calibration improves probability quality; linear models are comparatively well-behaved, while tree ensembles retain slightly higher ECE/Brier. (3) Shortcut probes and subgroup analyses suggest residual failure modes at extreme lengths and specific clusters; these are natural targets for future structure-aware or family-aware methods.

\section{Implementation Details and Reproducibility}

\textbf{CPU-only stack.} All models are implemented with scikit-learn and run on CPU. Feature extraction uses Biopython’s \texttt{ProtParam}. Homology clustering uses CD-HIT (we used 4.8.1 in our reference run). The reference notebook (e.g., Colab) contains package installs; no GPUs or proprietary software are required.

\textbf{Determinism.} We fix the Python/NumPy RNG seed to 1337 for splitting, model fitting, and bootstrap resampling. Composition-preserving shuffles use per-accession deterministic seeds so shuffled probes are reproducible across runs.

\textbf{Bootstrap details.} We use stratified bootstrap resampling ($n=200$) to ensure both classes are represented in each resample. For each bootstrap iteration, we independently sample with replacement from the positive and negative test examples, maintaining the original class balance. We report 2.5/97.5 percentiles as 95\% CIs.

\textbf{Operating-point computation.} On the empirical ROC $(\mathrm{FPR}(t),\mathrm{TPR}(t))$, we select the left-most threshold $t$ such that $\mathrm{FPR}(t)\geq 0.01$ for TPR@1\%FPR (and analogously the left-most threshold with $\mathrm{TPR}(t)\geq 0.95$ for FPR@95\%TPR).

\textbf{Calibration curves.} Reliability plots use 15 equal-width probability bins with mean predicted probability on the x-axis and empirical fraction positive on the y-axis; ECE averages bin-wise absolute gaps weighted by bin mass.

\textbf{Safety checks.} All artifacts (tables, plots) are aggregate and do not reveal raw sequences. The released CSV is metadata-only (accessions, cluster IDs, split labels, etc.).

\section{Ethics and Safety Statement}

This work is intended to \emph{improve} defensive capabilities by providing a transparent, reproducible baseline for screening, not to enable misuse. We release metadata only, avoid any design or release of novel sequences, and emphasize homology-aware evaluation that better reflects realistic risk surfaces. Our analyses, figures, and tables aggregate results without exposing hazardous sequences, aligning with the safety posture articulated by the SafeProtein initiative \citep{Fan2025SafeProtein} and biosecurity commentaries \citep{Hunter2024Biosecurity}.

\section{Conclusion}

We presented SafeBench-Seq, a tractable, CPU-only baseline and metadata-only benchmark for protein hazard screening that foregrounds homology-aware evaluation, calibrated probabilities, and shortcut probes. By clustering at ≤40\% identity and holding out entire clusters, we approximate “never-before-seen” homology groups and quantify discrimination at screening-relevant operating points with calibrated confidence. Empirically, random splits overestimate robustness relative to cluster holdouts; post-hoc calibration improves probability quality (Brier/ECE), particularly for linear models. Limitations include: (i) no explicit signal-peptide covariate (left for future work), (ii) sequence-only features (no structure/context), and (iii) moderate dataset size. Future directions include integrating lightweight structure proxies, explicit SP/N-terminal heuristics, and defense-in-depth (e.g., calibrated ensembling and risk-aware thresholds) while maintaining the same safety posture (metadata-only release). We hope SafeBench-Seq serves as a transparent reference point for evaluating sequence-level biohazard detectors.

\section{Limitations} 

Our approach has several constraints: (i) moderate dataset size (854 sequences) may limit statistical power for rare toxin families; (ii) sequence-only features ignore 3D structural information; (iii) we do not explicitly model signal peptides or subcellular localization signals that may be relevant for toxin function; (iv) performance at extreme operating points shows wide confidence intervals, reflecting the challenges of tail-event prediction in imbalanced settings.

\end{document}